\documentclass[conference]{IEEEtran}
\IEEEoverridecommandlockouts
\usepackage{cite}
\usepackage{bm}
\usepackage{amsmath,amssymb,amsfonts}
\usepackage{algorithm}
\usepackage{algorithmic}
\usepackage{graphicx}
\usepackage{subfigure}  
\usepackage{textcomp}
\usepackage{xcolor}
\usepackage{booktabs}
\usepackage{boldline}
\usepackage[amsmath]{ntheorem}
\usepackage{url}
\usepackage{color}

\allowdisplaybreaks[4]
\def\BibTeX{{\rm B\kern-.05em{\sc i\kern-.025em b}\kern-.08em
		T\kern-.1667em\lower.7ex\hbox{E}\kern-.125emX}}

\begin{document}

	\title{GAN-powered Deep Distributional Reinforcement Learning for Resource Management in Network Slicing}
	\author{\IEEEauthorblockN{Yuxiu Hua, Rongpeng Li, Zhifeng Zhao, Xianfu Chen, and Honggang Zhang}
		
	\thanks{%Manuscript received June 20, 2019; revised October 15, 2019; accepted November 6, 2019. This work was supported in part by National Key R\&D Program of China (No. 2017YFB1301003), National Natural Science Foundation of China (No. 61701439, 61731002), Zhejiang Key Research and Development Plan (No. 2019C01002, 2019C03131), the Project sponsored by Zhejiang Lab (2019LC0AB01), Zhejiang Provincial Natural Science Foundation of China (No. LY20F010016), the Fundamental Research Funds for the Central Universities (No. 2019QNA5010). (\emph{Corresponding author: Rongpeng Li}.)
		
		Yuxiu Hua, Rongpeng Li and Honggang Zhang are with the College of Information Science and Electronic Engineering, Zhejiang University, Hangzhou 310027, China (e-mail: \{21631087, lirongpeng, honggangzhang\}@zju.edu.cn).
		
		Zhifeng Zhao is with Zhejiang Lab, Hangzhou, China as well as the College of Information Science and Electronic Engineering, Zhejiang University, Hangzhou 310027, China (e-mail: zhaozf@zhejianglab.com).
		
		Xianfu Chen is with the VTT Technical Research Centre of Finland, Oulu 90570, Finland (e-mail: xianfu.chen@vtt.fi).}}
	
	\maketitle
	
	\begin{abstract}
		Network slicing is a key technology in 5G communications system. Its purpose is to dynamically and efficiently allocate resources for diversified services with distinct requirements over a common underlying physical infrastructure. Therein, demand-aware resource allocation is of significant importance to network slicing. In this paper, we consider a scenario that contains several slices in a radio access network with base stations that share the same physical resources (e.g., bandwidth or slots). We leverage deep reinforcement learning (DRL) to solve this problem by considering the varying service demands as the environment \emph{state} and the allocated resources as the environment \emph{action}. In order to reduce the effects of the annoying randomness and noise embedded in the received service level agreement (SLA) satisfaction ratio (SSR) and spectrum efficiency (SE), we primarily propose generative adversarial network-powered deep distributional Q network (GAN-DDQN) to learn the action-value distribution driven by minimizing the discrepancy between the estimated action-value distribution and the target action-value distribution. We put forward a reward-clipping mechanism to stabilize GAN-DDQN training against the effects of widely-spanning utility values.
		%For the sake of protecting the stability of GAN-DDQN's training process from the widely-spanning utility values, we put forward a reward-clipping mechanism. 
		Moreover, we further develop Dueling GAN-DDQN, which uses a specially designed dueling generator, to learn the action-value distribution by estimating the state-value distribution and the action advantage function. Finally, we verify the performance of the proposed GAN-DDQN and Dueling GAN-DDQN algorithms through extensive simulations.
	\end{abstract}
	
	
	\section{Introduction}

	The emerging fifth-generation (5G) mobile systems, armed with novel network architecture and emerging technologies, are expected to offer support for a plethora of network services with diverse performance requirements \cite{katsalis2017network}, \cite{li2017intelligent}. 
	Specifically, it is envisioned that 5G systems cater to a wide range of services differing in their requirements and types of devices, and going beyond the traditional human-type communications to include various kinds of machine-type communications \cite{foukas2017network}. According to ITU-R recommendations, it is a consensus that the key technologies of 5G wireless systems will spawn three generic application scenarios: enhanced mobile broadband (eMBB), massive machine-type communications (mMTC), and ultra-reliable and low-latency communications (URLLC) \cite{wp5d2017minimum}. Specifically, (a) eMBB supports data-driven use cases requiring high data rates across a wide coverage area; (b) mMTC supports a very large number of devices in a broad area, which may only send data sporadically, such as Internet of Things (IoT) use cases; (c) URLLC supports strict requirements on latency and reliability for mission-critical communications, such as remote surgery, autonomous vehicles or Tactile Internet. Serving a diverse set of use cases over the same network will increase complexity, which must be managed to ensure  acceptable service levels. As the services related to different use cases can have very different characteristics, they will impose varying requirements on the network resources. For example, considering mMTC devices for utility metering and parking sensors, they hardly move and do not require mobility management, i.e., the need to track location. On the other hand, sensors related to freight management would commonly move even across countries’ borders and would require mobility management, including roaming agreements.
	
	However, legacy mobile networks are mostly designed to provide services for mobile broadband consumers and merely consist of a few adjustable parameters like priority and quality of service (QoS) for dedicated services. Thus it is difficult for mobile operators to extend their networks into these emerging vertical services because of the different service requirements for network design and development. The concept of network slicing has recently been proposed to address this challenging problem; the physical and computational resources of the network can be sliced to meet the diverse needs of a range of 5G users \cite{katsalis2017network,zhou2016network,li2017network}. In this way, heterogeneous requirements can be served cost-effectively by the same physical infrastructure, since different network slice (NS) instances can be orchestrated and configured according to the specific requirements of the slice tenants.

	As a non-nascent concept, network slicing can be traced back to the Infrastructure as a Service (IaaS) cloud computing model \cite{afolabi2018network}, whereby different tenants share computing, networking and storage resources to create different isolated fully-functional virtual networks on a common infrastructure. In the context of 5G and beyond, network functions virtualization (NFV) and software defined networking (SDN) technologies serve as a basis for the core network slicing by allowing both physical and virtual resources to be used to provide certain services, enabling 5G networks to deliver different kinds of services to various customers \cite{zhang2017network}, \cite{ordonez2017network}. On the other hand, the Next Generation Mobile Networks (NGMN) alliance puts forward an evolved end-to-end network slicing idea while the Third-Generation Partnership Project (3GPP) also suggests that radio access network (RAN) should not be excluded to ``design specific functionality to support multiple slices or even partition of resources for different network slices" \cite{da2016impact,3gpptr22891}.
	
	However, in order to provide better-performing and cost-efficient services, RAN slicing involves more challenging technical issues for the realtime resource management on existing slices, since (a) for RANs, spectrum is a scarce resource and it is essential to guarantee the spectrum efficiency (SE) \cite{zhang2017network}; (b) the service level agreements (SLAs) with slice tenants usually impose stringent requirements; and (c) the actual demand of each slice heavily depends on the request patterns of mobile users (MUs) \cite{li2018deep}. Therefore, the classical dedicated resource allocation fails to address these problems simultaneously \cite{da2016impact}. Instead, it is necessary to intelligently allocate the radio resources (e.g., bandwidth or slots) to slices according to the dynamics of service requests from mobile users coherently \cite{vassilaras2017algorithmic} with the goal of meeting SLA requirements in each slice, but at the cost of acceptable SE. In this regard, there have been extensive efforts \cite{han2018slice,li2018deep,vo2018slicing,sun2018user,jiang2016network,doro2018low,zhou2018bandwidth,costanzo2018anetwork}. \cite{han2018slice} proposed an online genetic slicing strategy optimizer for inter-slice resource management. However, \cite{han2018slice} did not consider the explicit relationship between the required resource and SLA on a slice, as one slice might require more resources given its more stringent SLA. 
	\cite{vo2018slicing} considered the problem of different types of resources (bandwidth, caching, backhaul capacities) being allocated to NS tenants based on user demands. The authors proposed mathematical solutions, but the optimization problem would become intractable when the simulation parameters are scaled up (e.g., increasing the number of NSs or the shareable resources). \cite{sun2018user} mathematically analyzed the joint optimization problem of access control and bandwidth allocation in the multi-base station (BS) multi-NS scenario. However, the solutions therein are based on the assumption that different users have the same fixed demand rate, which is a condition unlikely to be found in practice. From the perspective of bandwidth usage-based pricing (UBP), \cite{zhou2018bandwidth} used game theory to analyze the relationship between Internet service providers (ISPs) and users, thereby improving the profit of ISPs and solving the peak-time congestion problem. However, the timeslot for bandwidth allocation in \cite{zhou2018bandwidth} was 1 hour, which is unrealistic in a situation where the number of demands changes drastically in a short period.
	
	In order to address the demand-aware resource allocation problem, one potential solution is reinforcement learning (RL). RL is an important type of machine learning where an agent learns how to perform optimal actions in an environment from observing state transitions and obtaining feedback (rewards/costs). In RL, the action value, $Q(s, a)$, describes the expected return, or the discounted sum of rewards, when performing action $a$ in state $s$. Usually, the $Q$ value can be estimated by classic value-based methods such as SARSA \cite{rummery1994line} and Q-learning \cite{watkins1992q} based on the Bellman equation. \cite{mnih2015human} used deep neural networks to approximate the $Q$ function, namely deep Q network (DQN), which demonstrated human-like performance on simple computer games and inspired a research wave of deep reinforcement learning (DRL). Besides modeling $Q(s, a)$, \cite{bellemare2017distributional} showed that we could learn the distribution of $Q(s, a)$ by the distributional analogue of Bellman equation; this approach improved the estimation of action values in an inherently randomness environment. Specifically, \cite{bellemare2017distributional} proposed C51 algorithm to minimize the Kullback–Leibler (KL) divergence between the approximated $Q$ distribution and the target $Q$ distribution calculated by the distributional Bellman optimality operator. Inspired by the theory of quantile regression \cite{koenker2001quantile}, \cite{dabney2018distributional} proposed the quantile regression DQN (QR-DQN) and thus successfully performed distributional RL over the Wasserstein metric, leading to the state-of-the-art performance. \cite{dabney2018implicit} extended QR-DQN from learning a discrete set of quantiles to learning the full quantile function and put forward the implicit Q network (IQN). Given the success of replacing $Q(s, a)$ by its distribution in \cite{bellemare2017distributional,dabney2018distributional,dabney2018implicit} as well as the reputation of generative adversarial network (GAN) for approximating distributions \cite{goodfellow2014generative}, it naturally raises a question of whether GAN is viable for approximating the action-value distribution and thus improving distributional RL.
	
	In the field of communications and networking, DRL has triggered tremendous research attention to solving resource allocation issues in some specific fields like power control \cite{li2018intelligentpower}, green communications \cite{xu2017adeep}, cloud RANs \cite{liu2017ahierarchical}, mobile edge computing and caching \cite{he2017software}. Given the challenging technical issues in resource management on existing NSs, the previous work in \cite{li2018deep} leveraged DQN to find the optimal resource allocation policy and investigated its performance. However, the method proposed in \cite{li2018deep} did not consider the effects of random noise on the calculation of SE and SLA satisfaction ratio (SSR). To mitigate the potential risk of incorrectly estimating the action value due to the randomness in SE and SSR, we intend to introduce the distributional RL to estimate the action-value distribution, thus avoiding the action-value overestimation or underestimation issue that plagues many traditional value-based RL algorithms (e.g., DQN). Meanwhile, the cutting-edge performance of Wasserstein generative adversarial network with gradient penalty (WGAN-GP) in the distribution approximation suggests to us that we might use it to learn the action-value distribution. To this end, we propose a new approach, the GAN-powered deep distributional Q network (GAN-DDQN), based on distributional RL and WGAN-GP, to realize dynamic and efficient resource allocation per slice. The main contributions of this paper are as follows:
	
	\begin{itemize}
	\item To find the optimal resource allocation policy under the uncertainty of slice service demands, we design the GAN-DDQN algorithm, where the generator network outputs a fixed number of particles that try to match the action-value distribution for each action. Such a design in GAN-DDQN can mitigate the effects of learning from a nonstationary environment and is significantly different from the concurrent yet independent works \cite{doan2018gan, freirich2019distributional} \footnote{\cite{doan2018gan} used a generator network that directly outputs action values and did not show any significant improvement of GAN Q-learning over conventional DRL methods. \cite{freirich2019distributional} used the policy iteration method \cite{sutton1999reinforcement} to loop through a two-step procedure for the value estimation and policy improvement, where GAN was only used to estimate the action-value distribution. Besides, \cite{freirich2019distributional} exploited a totally different framework without the target generator, which is a key component for GAN-DDQN.}.
	\item We demonstrate that the widely-spanning system utility values could destabilize GAN-DDQN's training process, and we correspondingly design a reward-clipping mechanism to reduce this negative impact. Specifically, we clip the system utility values to some constant values according to a straightforward rule with several heuristics-guided adjustable thresholds, and then use these constants as the final rewards in RL.
	\item GAN-DDQN suffers from the challenge that only a small part of the generator output is included in the calculation of the loss function during the training. To compensate for this, we further propose Dueling GAN-DDQN, which is a special solution derived from Dueling DQN \cite{wang2015dueling} and the discrete normalized advantage functions (DNAF) algorithm \cite{qi2019deep}. Dueling GAN-DDQN separates the state-value distribution from the action-value distribution and combines the action advantage function to obtain the action values. In addition, we elaborate on twofold loss functions that further take advantage of the temporal difference (TD) error information to achieve performance gains. The introduction of dueling networks to GAN-DDQN makes the work described in this paper significantly different from our previous work presented in IEEE Globecom 2019 \cite{hua2019gan}.
	\item Finally, we perform extensive simulations to demonstrate the superior efficiency of the proposed solutions over the classical methods, such as DQN, and provide insightful numerical results for the implementation details.
	\end{itemize}
	
	The remainder of the paper is organized as follows: Section II talks about some necessary mathematical backgrounds and formulates the system model. Section III gives the details of the GAN-DDQN, while Section IV presents the detailed simulation results. Finally, Section V summarizes the paper and offers prospects.
	
	\begin{table}[h!]
		\label{notations}
		\begin{center}
			\caption{Notations Used in This Paper}
			\begin{tabular}{l l}
				\toprule[1pt]
				Notation           & Definition                            \\ \midrule[1pt]
				$\mathcal{S}$      & State space                           \\
				$\mathcal{A}$      & Action space                          \\
				$P$                & Transition probability                \\
				$V$                & State-value function                  \\
				$Q$                & Action-value function           \\
  				$Z_v$              & Random variable to statistically model the state values \\
				$Z_q$              & Random variable to statistically model the action values \\
				$\mathcal{T}^{*}$  & Bellman optimality operator           \\
				$s$, $s^\prime$    & States                    \\
				$a$                & An action                      \\
				$r$                & A reward								\\
				$S_t$              & State at time $t$                  \\
				$A_t$              & Action at time $t$                    \\
				$R_t$              & Reward at time $t$                     \\
				$\gamma$           & Discount factor                       \\
				$\pi$              & Policy                                \\
				$J$				   & System utility                       \\
				$\alpha$           & Weight of the SE                      \\
				$\bm{\beta}$       & Weight of the SSR                     \\
				$\tau$             & Quantile samples                      \\
				$\lambda$          & Gradient penalty coefficient          \\
				$n_{critic}$	   & The number of discriminator updates per training iteration \\
				\midrule[1pt]
			\end{tabular}
		\end{center}
	\end{table}
	
	\section{Preliminaries and System Model}
	\subsection{Preliminaries}
	Table \ref{notations} lists the important notations used in this paper.
	An agent tries to find the optimal behavior in a given setting through interaction with the environment, which can be treated as solving an RL problem. This interactive process can be modeled as a Markov Decision Process $(\mathcal{S}, \mathcal{A}, R, P, \gamma)$, where $\mathcal{S}$ and $\mathcal{A}$ denote the state and action spaces, $R$ is the reward, $P(\cdot|s, a)$ is the transition probability, and $\gamma \in (0,1]$ is a discount factor. A policy $\pi(\cdot|s)$ maps a state to a distribution over actions. The state-value function of a state $s$ under a policy $\pi(\cdot|s)$, denoted $V^\pi(s)$, is the expected return when starting in $s$ and following $\pi$ thereafter. Similarly, we define the value of taking action $a$ in state $s$ under the policy $\pi$, denoted $Q^\pi(s,a)$, as the expected return starting from $s$, taking the action $a$, and thereafter following policy $\pi$. Mathematically, the state-value function is
	\begin{equation}
	\label{state-value-func}
		V^{\pi}(s)=\mathbb{E}_{\pi, P}\left[\sum_{t=0}^{\infty} \gamma^{t}R_t | S_{0}=s\right],
	\end{equation}
	and the action-value function is 
	\begin{equation}
	\label{action-value-func}
		Q^{\pi}(s, a)=\mathbb{E}_{\pi, P}\left[\sum_{t=0}^{\infty} \gamma^{t} R_t | S_{0}=s, A_{0}=a\right],
	\end{equation}
	where $\mathbb{E}$ denotes the expectation. The relationship between the value of a state and the values of its successor states is expressed by the Bellman equation for $V^\pi$
	\begin{equation}
	\label{bellman-V}
		V^{\pi}(s)=\mathbb{E}_{\pi, P}\left[R+\gamma V^{\pi}(s^{\prime})\right].
	\end{equation}
	Similarly, the Bellman equation for $Q^\pi$ is
	\begin{equation}
	\label{bellman-Q}
	Q^{\pi}(s, a)=\mathbb{E}_{\pi, P}\left[R+\gamma Q^{\pi}\left(s^{\prime}, a^{\prime}\right)\right],
	\end{equation}
	where $s^{\prime}$ and $a^{\prime}$ can be derived from the transition probability $P(\cdot|s, a)$ and a policy $\pi(\cdot|s^{\prime})$, respectively.
	
	The goal of RL is to find the optimal policy which yields the maximum $Q(s, a)$ for all $s$ and $a$. Let $\pi^{*}=\arg \max _{\pi}Q^{\pi}(s,a)$ be the optimal policy and let $Q^{*}(s, a)$ be the corresponding action-value function. $Q^{*}(s, a)$ satisfies the following Bellman optimality equation
	\begin{equation}
	\label{opt-bellman-Q}
	Q^{*}\left(s, a\right)=\mathbb{E}_{\pi^{*}, P}\left[R+\gamma \max _{a^{\prime} \in {\mathcal{A}}} Q^{*}\left(s^{\prime}, a^{\prime}\right)\right].
	\end{equation}
	Eq. \eqref{opt-bellman-Q} illustrates the temporal consistency of the action-value function, which allows for the design of learning algorithms. Define the Bellman optimality operator $\mathcal{T}^{*}$ as
	\begin{equation}
	\mathcal{T}^{*} Q\left(s, a\right) =\mathbb{E}_{\pi, P}\left[R+\gamma \max _{a^{\prime} \in {\mathcal{A}}} Q\left(s^{\prime}, a^{\prime}\right)\right].
	\end{equation}
	When $\gamma \in (0,1)$, starting from any $Q_t(s, a)$, iteratively applying the operator $Q_{t+1}(s, a) \leftarrow \mathcal{T}^{*} Q_{t}(s, a)$ leads to convergence $Q_{t}(s, a) \rightarrow Q^{*}(s, a)$ as $t \rightarrow \infty$ \cite{sutton1999reinforcement}.
	
	In high dimensional cases, it is critical to use function approximation as a compact representation of action values. Let $Q_{\theta}(s,a)$ denote a function with parameter $\theta$ that approximates a table of action values with entry $(s,a)$. The optimization aim is to find $\theta$ such that $Q_{\theta}(s,a) \approx Q^{*}(s,a)$, and the optimal solution can be found by iteratively leveraging the Bellman optimality operator $\mathcal{T}^{*}$. In other words, the optimal parameter $\theta$ can be approached by minimizing the squared TD error
	\begin{equation}
	\label{eq2}
	\zeta^{2}=\left[r+\gamma \max _{a^{\prime} \in {\mathcal{A}}} Q_{\theta}\left(s^{\prime}, a^{\prime}\right)-Q_{\theta}\left(s, a\right)\right]^{2}
	\end{equation}  
	over samples $(s, a, r, s^{\prime})$, which are randomly selected from a replay buffer \cite{mnih2013playing} that stores transitions which record the interaction between an agent and the environment when following the policy driven by $Q_{\theta}$. In cases where $Q_{\theta}(s,a)$ is linear, the iterative process to find $Q^{*}(s,a)$ can be shown to converge \cite{Tsitsiklis1996}. However, in cases where $Q_{\theta}(s,a)$ is nonlinear (e.g., a neural network), $Q_{\theta}(s,a)$ becomes more expressive at the cost of no convergence guarantee. A number of DRL algorithms are designed following the above formulation, such as DQN \cite{mnih2015human} and Dueling DQN \cite{wang2015dueling}. 
	
	\begin{figure*}
		\centering
		\includegraphics[width=0.8\linewidth]{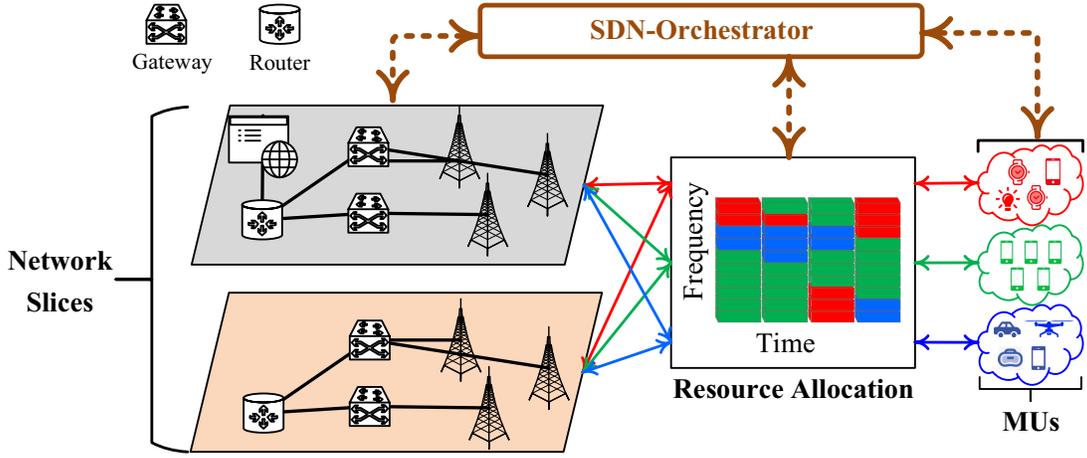}
		\caption{The considered scenario showing uplink and downlink transmissions on different NSs.}
		\label{fig:system_model}
	\end{figure*}

	\subsubsection{Distributional Reinforcement Learning}
	The main idea of distributional RL \cite{bellemare2017distributional} is to work directly with the distribution of returns rather than their expectation (i.e., $Q^{\pi}$), so as to increase robustness to hyperparameter variation and environment noise \cite{barth2018distributed}. Let the random variable $Z_q^{\pi }(s, a)$ be the return obtained by following a policy $\pi$ to perform action $a$ from the state $s$. Notably, the value of $Z_q^{\pi }(s, a)$ varies due to the unexpected randomness in the environment. Then we have 
	\begin{equation}
	\label{zandq}
	Q^{\pi}(s, a)=\mathbb{E}\left[Z_q^{\pi}(s, a)\right],
	\end{equation}
	 and an analogous distributional Bellman equation, that is,
	\begin{equation}
	\label{belleqq}
	Z_q^{\pi}(s, a) \stackrel{D}{=} R+\gamma Z_q^{\pi}\left(s^{\prime}, a^{\prime}\right),
	\end{equation}
	where $A \stackrel{D}{=} B$ denotes that random variable $A$ has the same probability law as $B$. Therefore, a distributional Bellman optimality operator $\mathcal{T}^{*}$ can be defined by
	\begin{equation}
	\label{optbellq}
	\mathcal{T}^{*} Z_q(s, a) \stackrel{D}{=} R+\gamma Z_q\left(s^{\prime}, \underset{a^{\prime} \in {\mathcal{A}}}{\arg \max } \mathbb{E} [Z_q\left(s^{\prime}, a^{\prime}\right)]\right).
	\end{equation}
		
	In traditional RL algorithms, we seek the optimal $Q$ function approximator by minimizing a scalar value $\zeta^{2}$ in Eq. \eqref{eq2}. In distributional RL, our objective is to minimize a statistical distance:
	\begin{equation}
	\label{eq5}
	\sup _{s, a} \operatorname{dist}\left(\mathcal{T}^{*} Z_q(s, a), Z_q(s, a)\right),
	\end{equation}
	where $\operatorname{dist}(A, B)$ denotes the distance between random variable $A$ and $B$, which can be measured by many metrics, such as KL divergence \cite{bellemare2017distributional}, $p$-Wasserstein \cite{dabney2018distributional}, etc.
	In \cite{bellemare2017distributional}, Bellemare \emph{et al.} proved that the distributional Bellman equation is a contraction in $p$-Wasserstein distance, but the distributional Bellman optimality operator is not necessarily a contraction, which provides a guideline for metric selection. C51 algorithm \cite{bellemare2017distributional} approximates the distribution over returns using a fixed set of equidistant points and optimizes Eq. \eqref{eq5} by minimizing KL divergence. Different from KL divergence based on the probability density function, $p$-Wasserstein metric builds on the cumulative distribution function. Assume that there are two real-valued random variables $U$ and $V$ with respective cumulative distribution functions $F_U$ and $F_V$, the $p$-Wasserstein between them is given by\footnote{We further explain the advantage of the Wasserstein metric in the next part.}
	\begin{equation}
	\label{p-Wasserstein}
	W_{p}(U, V)=\left(\int_{0}^{1}\left|F_{U}^{-1}(\omega)-F_{V}^{-1}(\omega)\right|^{p} d \omega\right)^{1 / p}.
	\end{equation}
	Theoretically, the distributional Bellman optimality operator is a strict contraction in $p$-Wasserstein distance; that is, minimizing Eq. \eqref{eq5} with $p$-Wasserstein distance can give the optimal action-value distribution. QR-DQN \cite{dabney2018distributional} used the values on some uniformly distributed quantiles to describe the action-value distribution, leveraging the loss of quantile regression to train the neural network, which is an effective approach to minimizing $1$-Wasserstein distance. Therefore, QR-DQN obtains a better balance between theory and practice by working on a special case (i.e., $1$-Wasserstein distance, the special case for $p$-Wasserstein distance with $p=1$).
	
	\subsubsection{Generative Adversarial Network}
	GAN \cite{goodfellow2014generative} is intended to learn the distribution of data from all domains, mostly image, music, text, etc., to generate convincing data. GAN consists of two neural networks, a generator network $G$ and a discriminator network $D$, which are engaged in a zero-sum game against each other. The network $G$ takes an input from a random distribution and maps it to the space of real data. The network $D$ obtains input data from both real data and the output of $G$, and attempts to distinguish the real data from the generated data. The two networks are trained by gradient descent algorithms in alternating steps. 
	
	The classical GAN minimizes Jensen-Shannon (JS) divergence between the real data and generated data distributions. However, \cite{arjovsky2017wasserstein} shows that JS metric is not continuous and does not provide a usable gradient all the time. To overcome this shortcoming, \cite{arjovsky2017wasserstein} proposed WGAN, in which the JS metric is replaced by $1$-Wasserstein distance that provides sufficient gradients almost everywhere. Given that the equation for $1$-Wasserstein distance is highly intractable, WGAN uses Kantorovich-Rubinstein duality to simplify the calculation, but introducing an essential constraint that ensures the discriminator is an appropriate $1$-Lipschitz function. WGAN satisfies the constraint by clipping the weights of the discriminator to be within a certain range that is governed by a hyperparameter. Furthermore, \cite{gulrajani2017improved} proposed WGAN-GP and adopted gradient penalty to enforce the $1$-Lipschitz constraint instead of simply clipping weights. Its optimization objective is formulated as follow:
	\begin{equation}
	\label{minmax}
	\min _{G} \max _{D \in \mathcal{D}} \mathbb{E}_{\mathbf{x} \sim p_{\text {data}}}[ D(\mathbf{x})]-\mathbb{E}_{\mathbf{z} \sim p_{\boldsymbol{z}}(\mathbf{z})}[D(G(\mathbf{z}))]+p(\lambda),
	\end{equation} 
	where $\mathcal{D}$ denotes the set of $1$-Lipschitz functions, $\mathbf{x}$ denotes the samples from real data, $\boldsymbol{z}$ denotes the samples from a random distribution, and $p(\lambda)=\frac{\lambda}{2} \left(\left\|\nabla_{\hat{\mathbf{x}}} D(\hat{\mathbf{x}})\right\|_{2}-1\right)^{2}$, $\hat{\mathbf{x}}=\varepsilon \mathbf{x}+(1-\varepsilon) G(\mathbf{z})$, $\varepsilon \sim U(0, 1)$. Gradient penalty increases the computational complexity but it does make WGAN-GP perform much better than previous GANs.
	
	\begin{figure*}
		\centering
		\includegraphics[width=0.65\linewidth]{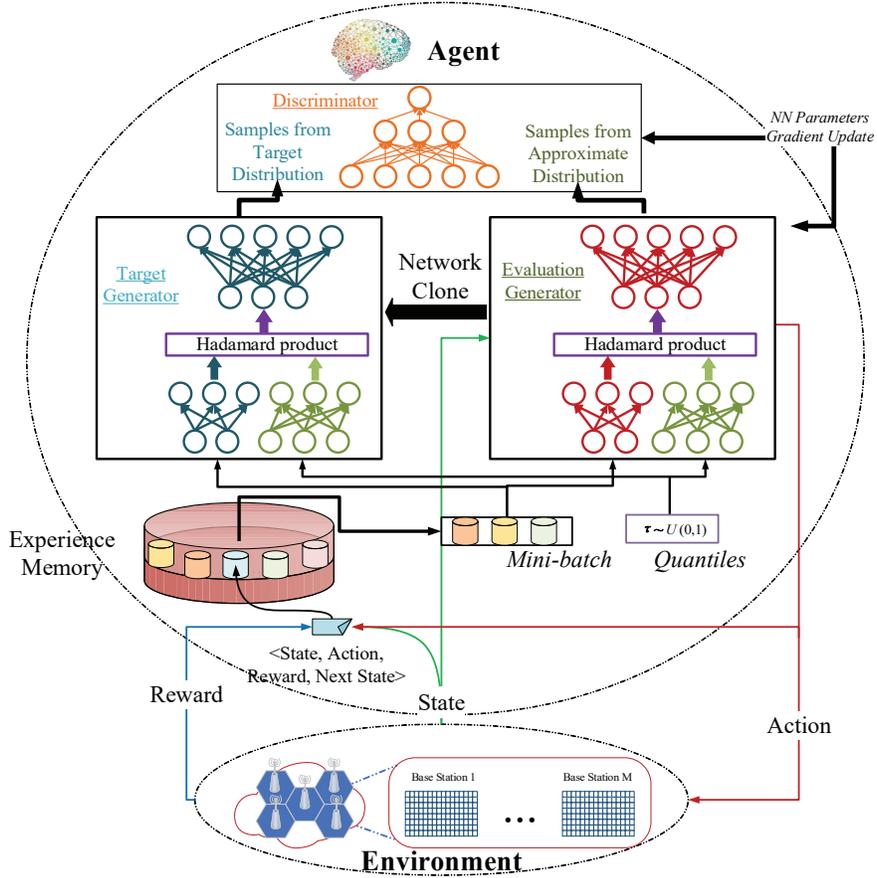}
		\caption{An illustration of GAN-DDQN for resource management in network slicing.}
		\label{fig:model}
	\end{figure*}

	\subsection{System Model} 

	Fig. \ref{fig:system_model} illustrates the SDN-based system model for dynamic allocation of wireless bandwidth in the RAN scenario with uplink and downlink transmissions. We consider the downlink case in this paper. In this respect, \cite{costanzo2018anetwork} built an SDN-based C-RAN testbed that realizes the dynamic allocation of radio resources (e.g., wireless bandwidth) by using frequency division duplex scheme, and the demonstration was presented in \cite{costanzo2018sdn}. Under the framework of hierarchical network slicing, we consider a RAN scenario with a single BS, where a set $\mathcal{N}$ of NSs share the aggregated bandwidth $W$\footnote{In fact, such a bandwidth allocation could be realized by physical multiplexing methods \cite{networkslcingarch}. Meanwhile, the shared resources could be temporal slots as well. However, for simplicity of representation, we take the bandwidth allocation problem as an example.}. The bandwidth is allocated to each NS according to the number of demands for the corresponding type of service. For an NS, say NS $n$, it provides a single service for a set of users $\mathcal{U}_n$. We consider a timeslot model where the slicing decision is updated according to the demand of users periodically (e.g., 1 second). In one timeslot, the number of demands NS $n$ receives is denoted as $d_n$, which partially determines the wireless bandwidth that the BS allocates to this NS, denoted as $w_n$.
	
	The objective of our work is to find an optimal bandwidth-allocation solution that maximizes the system utility, denoted by $J$, which can be described by the weighted sum of SE and SSR. We now study the two sub-objectives, respectively. Let $r_{u_n}$ be the downlink rate of user $u_n$ served by NS $n$, which is, for simplicity, defined by Shannon theory as follows
	\begin{equation}
	r_{u_n}=w_{n}\text{log}(1+\text{SNR}_{u_n}), \forall u_n \in \mathcal{U}_n,
	\end{equation}
	where $\text{SNR}_{u_n}$ is the signal-to-noise-ratio between user $u_n$ and the BS. $\text{SNR}_{u_n}$ can be given as
	\begin{equation}
	\text{SNR}_{u_n}=\frac{g_{u_n}P_{u_n}}{N_0w_n},
	\end{equation}
	where $g_{u_n}$ is the average channel gain that captures path loss and shadowing from the BS to the user $u_n$, $P_{u_n}$ is the transmission power, and $N_0$ is the single-side noise spectral density. Given the transmission rate, SE can be defined as follows
	\begin{equation}
	\text{SE}=\frac{\sum_{n\in\mathcal{N}}\sum_{u_n\in\mathcal{U}_n}r_{u_n}}{W}.
	\end{equation}

	On the other hand, SSR of NS $n$ is obtained by dividing the number of successfully transmitted packets by the total number of arrived packets on NS $n$. Before formulating this problem, we define $\mathcal{Q}_{u_n}$ as the set of packets sent from the BS to user $u_n$, determined by the actual traffic demand patterns, and define a binary variable $x_{q_{u_n}} \in 0,1$, where $x_{q_{u_n}}=1$ indicates that the packet $q_{u_n} \in \mathcal{Q}_{u_n}$ is successfully received by user $u_n$, i.e., the downlink data rate $r_{u_n}$ and the latency $l_{q_{u_n}}$ are simultaneously satisfied.
	Therefore, $x_{q_{u_n}}=1$ if and only if $r_{u_n}\geq\overline{r}_n$ and $l_{q_{u_n}}\leq\overline{l}_n$, where $l_{q_{u_n}}$ denotes the latency that takes account of both queuing delay and transmission delay. $\overline{r}_n$ and $\overline{l}_n$ are the predetermined rate and latency values according to the SLA for service type $n$. 
	We can formulate the SSR for NS $n$ as:	
	\begin{equation}
	\text{SSR}_n=\frac{\sum_{u_n\in\mathcal{U}_n}\sum_{q_{u_n}\in\mathcal{Q}_{u_n}}x_{q_{u_n}}}{\sum_{u_n\in\mathcal{U}_n}\left|\mathcal{Q}_{u_n}\right|},
	\end{equation}
	where $\left|\mathcal{Q}_{u_n}\right|$ denotes the number of packets sent from the BS to user $u_n$.

	The bandwidth allocation problem in the RAN network slicing is formulated as follows
	\begin{align}
	\label{optimization}
	\begin{split}
	\max_{w_n} \quad & \alpha\text{SE} + \sum_{n\in\mathcal{N}}\beta_n \cdot \text{SSR}_n \\
	=\max_{w_n} \quad & \alpha\frac{\sum_{n\in\mathcal{N}}\sum_{u_n\in\mathcal{U}_n}r_{u_n}}{W}\\ &+ \sum_{n\in\mathcal{N}}\beta_n \cdot \frac{\sum_{u_n\in\mathcal{U}_n}\sum_{q_{u_n}\in\mathcal{Q}_{u_n}}x_{q_{u_n}}}{\sum_{u_n\in\mathcal{U}_n}\left|\mathcal{Q}_{u_n}\right|},
	\end{split}\\
	\label{constraint_bandwidth}
	\textrm{s.t.} \quad & \sum_{n\in\mathcal{N}}w_n=W,\\
	\label{constraint_demand}
	& \sum_{u_n\in\mathcal{U}_n}\left|\mathcal{Q}_{u_n}\right|=d_n,\\
	\label{constraint_int}
	& x_{p_{u_n}}=
	\begin{cases} 
		1, & r_{u_n}\geq\overline{r}_n \quad\&\quad l_{p_{u_n}}\leq\overline{l}_n, \\
		0, & otherwise.
	\end{cases}
	\end{align}
	where $\alpha$ and $\bm{\beta}=[\beta_1, \beta_2, \cdots \beta_n]$ are the coefficients that adjust the importance of SE and SSR, and $\beta_n$ refers to the importance weight of $\text{SSR}_n$. In this problem, the objective is to maximize two components: (a) the spectral efficiency (i.e., SE), and (b) the proportion of the packets satisfying the constraint of data rate and latency (i.e., SSR).
	
	Notably, in the current timeslot, $d_n$ depends on both the number of demands and the bandwidth-allocation solution in the previous timeslot, since the maximum transmission capacity of RAN belonging to one service is tangled with the provisioning capabilities for this service. For example, the TCP sending window size is influenced by the estimated channel throughput. Therefore, the traffic demand varies without knowing a prior transition probability, making Eq. \eqref{optimization} difficult to yield a direct solution. However, RL promises to be applicable to tackle this kind of problem. Therefore, we refer to RL to find the optimal policy for network slicing. In particular, consistent with \cite{li2018deep}, we map the RAN scenario to the context of RL by taking the number of arrived packets in each slice within a specific time window as the state, and the bandwidth allocated to each slice as the action.

	\section{GAN-powered Deep Distributional Q Network}
	In this section, we describe the proposed GAN-DDQN algorithm, shown in Fig. \ref{fig:model}, that address the demand-aware resource allocation problem in network slicing. We then discuss the methods of improving the performance of the algorithm and analyze its convergence.
		\begin{algorithm}
		\label{algorithm}
		\caption{GAN-DDQN}
		\begin{algorithmic}[1]
			\STATE Initialize a generator $G$ and a discriminator $D$ with random weights $\bm{\theta}_G$ and $\bm{\theta}_D$ respectively, the number of particles $N$, gradient penalty coefficient $\lambda$, batch size $m$, discount factor $\gamma$.
			\STATE Initialize a target generator $\hat{G}$ with weight $\bm{\theta}_{\hat{G}} \gets \bm{\theta}_G$, a replay buffer $\mathcal{B} \gets \varnothing$, the iteration index $t=0$.
			\REPEAT
			\STATE The agent observes $S_t=s$.
			\STATE The agent samples $\tau \sim U(0, 1)$.
			\STATE The agent calculates $Q(s, a) = \dfrac{1}{N} \sum G^{(a)}\left(s, \tau\right), \forall a\in \mathcal{A}$.
			\STATE The agent performs $a^* \gets \arg\max_{a} Q(s, a), \forall a\in \mathcal{A}$.
			\STATE The agent receives the system utility $J$ and observes $S_{t+1}=s^\prime$.
			\STATE The agent performs the reward-clipping with respect to $J$ and gets the reward $r$.
			\STATE The agent stores transition $(s, a^*, r, s^\prime)$ in $\mathcal{B}$.
			\STATE If $\mathcal{B}$ is full, the agent updates the weights of network $G$ and network $D$ every $K$ iterations.
			\STATE \# \emph{Train GAN}
			\REPEAT
			\STATE The agent samples a minibatch $\left\{s, a, r, s^{\prime}\right\}_{i=1}^{m}$ from $\mathcal{B}$ without replacement.
			\STATE The agent samples a minibatch $\{\tau\}_{i=1}^{m} \sim U(0,1)$.
			\STATE The agent gets the target action-value particles ${y}_i = r_i + \gamma \hat{G}^{(a^*_i)}(s^{\prime}_i, \tau_i)$, where the optimal action is $a^*_i = \arg\max_{a} \dfrac{1}{N} \sum \hat{G}^{(a)}\left(s^{\prime}_i, \tau_i\right), \forall a\in \mathcal{A}$.
			\STATE The agent samples a minibatch $\{\varepsilon\}_{i=1}^{m} \sim U(0,1)$, and sets $\hat{{x}}_i=\varepsilon_i {y}_i + (1-\varepsilon_i) G^{(a_i)}(s_i, \tau_i)$.
			\STATE The agent updates the weights $\bm{\theta}_D$ by leveraging gradient descent algorithm to $\frac{1}{m} \sum^{m}_{i=1}\mathcal{L}_i$, where $\mathcal{L}_i=D(G^{(a_i)}(s_i, \tau_i)) - D(y_i) + \lambda \left(\left\|\nabla_{\hat{{x}}_i} D(\hat{{x}}_i)\right\|_{2}-1\right)^{2}$.
			\STATE The agent updates the weights $\bm{\theta}_G$ by leveraging gradient descent algorithm to $-\frac{1}{m} \sum^{m}_{i=1} D(G^{(a_i)}(s_i, \tau_i))$
			\UNTIL All the transitions in $\mathcal{B}$ are used for training.
			\STATE The agent clones network $G$ to the target network $\hat{G}$ every $C$ iterations by resetting $\bm{\theta}_{\hat{G}} = \bm{\theta}_G$.
			\STATE The iteration index is updated by $t \gets {t+1} $.
			\UNTIL A predefined stopping condition (e.g., the $\frac{1}{m} \sum^{m}_{i=1}\mathcal{L}_i$, the preset number of iterations, etc.) is satisfied.		
		\end{algorithmic}
	\end{algorithm}

	\subsection{GAN-DDQN Algorithm}
	Our previous work \cite{li2018deep} has discussed how to apply RL to the resource slicing problem. 
	However, the DQN algorithm used in that work is based on the expectation of the action-value distribution, and thus does not take into account the adverse effects of random noise on the received values of SE and SSR. To overcome this problem, we resort to the combination of the distributional RL and GAN. In this regard, we introduce WGAN-GP to learn the optimal action-value distribution. Specifically, the generator network $G$ outputs a fixed number of samples (we refer to them as \textit{particles} for clarity) that characterize the estimated action-value distribution learned by network $G$. Similar to \cite{mnih2015human}, we leverage a target generator network $\hat{G}$ to obtain the target action-value particles. The discriminator network $D$ realizes the 1-Wasserstein criterion when it attempts to minimize the distance between the estimated action-value particles and the target action-value particles calculated by the Bellman optimality operator. GAN-DDQN is able to approximate the optimal action-value distribution by alternately updating networks $G$ and $D$. 

\begin{figure*}
		\centering
		\subfigure[An illustration of GAN-DDQN algorithm.]{\includegraphics[width=0.65\linewidth]{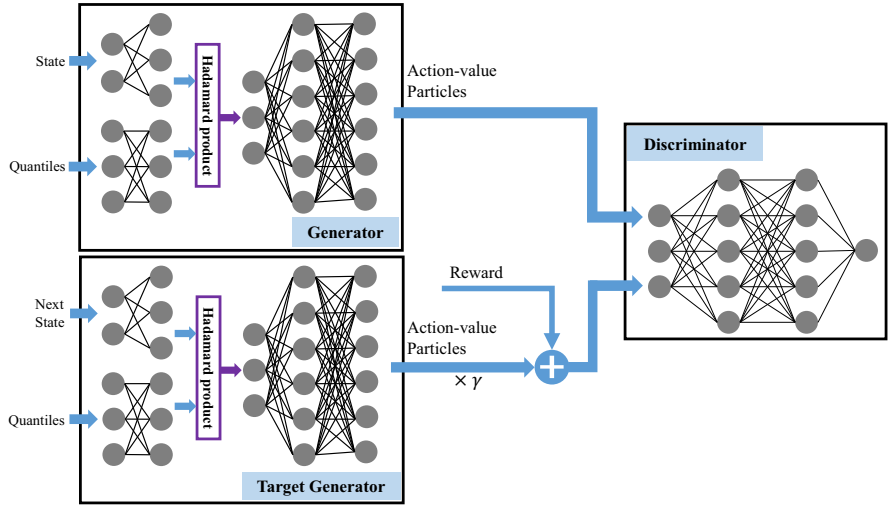}} \\
		\subfigure[An illustration of Dueling GAN-DDQN algorithm.]{\includegraphics[width=0.7\linewidth]{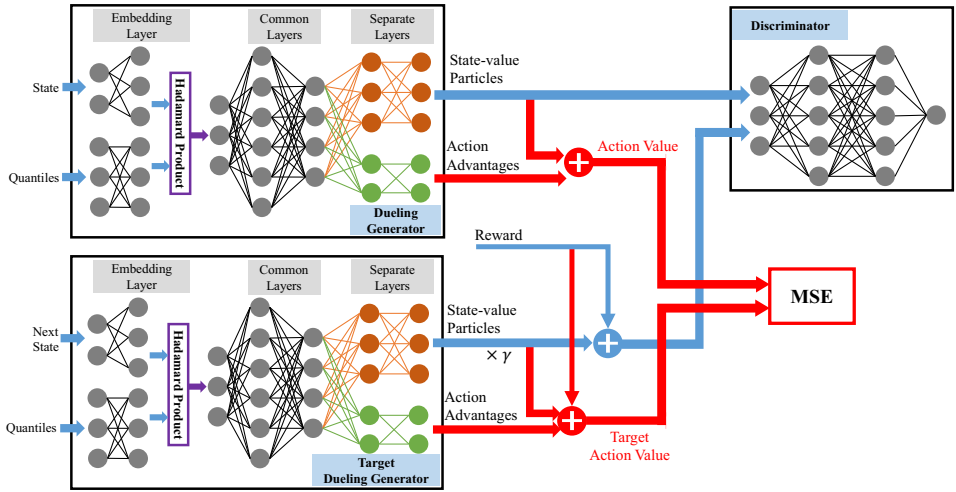}}
		\caption{The comparison of GAN-DDQN and Dueling GAN-DDQN.}
		\label{fig:duelingGANDDQN}
	\end{figure*}

	Before we introduce the details of GAN-DDQN algorithm, it is necessary to describe the structure of the networks $G$ and $D$. Network $G$ consists of three components, which are responsible for state embedding, sample embedding, and particles generation.
	%can be regarded as the integration of three components, namely state-embedding part, sample-embedding part and, generation part, respectively. 
	The state-embedding and sample-embedding components are both built with two neural layers and process the input state and the quantile samples in parallel. Then, the output of these two components are combined through Hadamard product operation. This step is consistent with \cite{dabney2018implicit} to force interaction between the state embedding and sample embedding. Afterwards, the particles generation component, which contains multiple neural layers, takes the fused information as input, outputting several sets of particles where each set is treated as a representation of the corresponding action-value distribution. On the other hand, network $D$ is a multilayer perceptron (MLP) with one neuron in the output layer. Fig. \ref{fig:duelingGANDDQN}(a) further details the structure of GAN-DDQN.

	The GAN-DDQN algorithm can be explained as follows, without loss of generality. At iteration $t$, the agent feeds the current state $S_t=s$ and the samples $\tau$ from a uniform distribution (e.g., $U(0, 1)$) to network $G$; $\tau$ is the quantile values of the action-value distribution \cite{dabney2018implicit}.
	Network $G$ outputs a set of estimated action-value particles, denoted as $G(s, \tau)$, where the particles belonging to action $a$ are denoted as $G^{(a)}(s, \tau)$; the number of $G^{(a)}(s, \tau)$ is $N$. Then, the agent calculates $Q(s, a)=\dfrac{1}{N} \sum G^{(a)}\left(s, \tau\right),\forall a\in \mathcal{A}$, and selects $a^*=\arg\max_{a} Q(s, a), \forall a\in \mathcal{A}$ to perform. As a result, the agent receives a reward $r$, and the environment moves to the next state $S_{t+1}=s^\prime$. The tuple $(s, a^*, r, s^\prime)$ is stored into the replay buffer $\mathcal{B}$. When $\mathcal{B}$ is full, the agent updates networks $G$ and $D$ using all the transition tuples in $\mathcal{B}$ every $K$ iterations.

	In the training and updating process, the agent first randomly selects $m$ transitions from $\mathcal{B}$ as a minibatch for training GAN-DDQN. Then, the agent executes the Bellman optimality operator on each transition of the selected minibatch and obtains the target action-value particles. For example, the target action-value particles for the transition $i$ is ${y}_i = r_i + \gamma \hat{G}^{(a^*_i)}(s^{\prime}_i, \tau_i)$ where $a^*_i$ is the action with the maximum expectation of action-value particles, i.e.,
	$a^*_i = \arg\max_{a} \dfrac{1}{N} \sum\hat{G}^{(a)}\left(s^{\prime}_i, \tau_i\right)$. Finally, the agent uses the following loss functions to train networks $D$ and $G$, respectively:
	\begin{align} 
	\begin{split}
	\mathcal{L}_D & = \underset{\underset{(s, a) \sim \mathcal{B}}{\tau \sim U(0, 1)}}{\mathbb{E}}[D(G^{(a)}(s, \tau))] -  \\	
	&\underset{(s, a, r, s^\prime) \sim \mathcal{B}}{\mathbb{E}}[ D(y)]+p(\lambda), 
	\end{split}\\
	\mathcal{L}_G &= -\underset{\underset{(s, a) \sim \mathcal{B}}{\tau \sim U(0, 1)}}{\mathbb{E}}[D(G^{(a)}(s, \tau))]
	\end{align}
	where $p(\lambda)$ is as mentioned in Eq. \eqref{minmax}. The training goal for network $D$ is to increase its accuracy in distinguishing the target action-value particles from the action-value particles produced by network $G$. The goal of training network $G$, on the other hand, is to improve its ability to generate the action-value particles that ``fool" network $D$ as much as possible. Note that in order to further stabilize the training process, we update the target network $\hat{G}$ every $C$ iterations. 
	
	Step by step, we incorporate the aforementioned methods and establish the GAN-DDQN as in Algorithm 1.
	
	\subsection{Convergence Analysis}

	It has been proven in \cite{bellemare2017distributional} that the distributional RL can converge when the metric for diverging distributions is $p$-Wasserstein distance. On the other hand, the fundamental guidance for distinguishing the target and estimated distributions in WGAN-GP is $1$-Wasserstein distance. Therefore, the convergence of GAN-DDQN can be analyzed from the perspective of WGAN-GP's convergence on the data sampled from the dynamic RL interaction process. However, as explored in \cite{mescheder2018gan}, in many currently popular GAN architectures, converging to the target distribution is not guaranteed and oscillatory behavior can be observed. This is a twofold challenge for GAN-DDQN, as we must ensure both the stationarity of the target distribution and the convergence of the WGAN-GP to this target distribution.
	
	In an idealized WGAN-GP, the generator should be able to learn from the target distribution, and the discriminator should be able to learn any $1$-Lipschitz function to produce the exact Wasserstein distance. However, the target distribution will not be stationary as the target network $\hat{G}$ regularly updates its weights; thus an idealized WGAN-GP might not be successful in practice. Fortunately, a slight change in the target distribution has little effect on the convergence of WGAN-GP. For example, suppose the real distribution that is the ultimate learning goal of WGAN-GP is a Gaussian distribution with a mean of 100 and a standard deviation of 1, and suppose that the target distribution that the WGAN-GP is expected to approximate at each stage is a Gaussian distribution with a standard deviation of 1 and a mean that starts at 0, increasing periodically by $\Delta \mu$. WGAN-GP will need more updates to learn the target distribution if $\Delta \mu$ is large; the number of updates is difficult to determine. However, if $\Delta \mu$ is small, a few times of updates is sufficient for WGAN-GP to learn the changed target distribution. Hence, the small $\Delta \mu$ is more potential to enable WGAN-GP to learn the real distribution smoothly.
	
	To analyze the convergence characteristic of WGAN-GP while avoiding directly dealing with sophisticated data and WGAN-GP model, \cite{mescheder2018gan} introduces a simple but illustrative model, namely Dirac-WGAN-GP. Specifically, Dirac-WGAN-GP consists of a linear discriminator $D_{\psi}(x)=\psi \cdot x$ and a generator with parameter $\theta$ that indicates the position of  the Dirac distribution (i.e., $\delta_{\theta}$) output by the generator. Whilst the real data distribution $p_d$ is given by a Dirac-distribution concentrated at $\xi$ (i.e., $\delta_{\xi}$). It is worthy to further investigate the training characteristic of Dirac-WGAN-GP when the real data distribution (i.e., $\delta_{\xi}$) is varying during the training process, like the typical situation in RL. Consistent with \cite{mescheder2018gan}, we carry out analysis based on Dirac-WGAN-GP, and we have the following Theorem \ref{therorem1}
	
	\theoremheaderfont{}
	\theoremseparator{:}
	\newtheorem*{theorem}{Theorem 1}
	\begin{theorem}\label{therorem1}
		\textit{When trained with gradient descent with a fixed number of the generator and the discriminator updates and a fixed learning rate $h>0$, if the value of $\xi$ varies dramatically, Dirac-WGAN-GP needs more learning steps to converge from the old optimal boundary to the new one after the variation of $\xi$.}
	\end{theorem}
	We leave the proof of Theorem \ref{therorem1} in Appendix. As for WGAN-GP, we further have the following Corollary \ref{corollary1}
	\newtheorem*{corollary}{Corollary 1}
	\begin{corollary}
		\label{corollary1}
		\textit{WGAN-GP could converge to the optimal boundary more rapidly if the real data change by a small amount.}
	\end{corollary}

	Corollary \ref{corollary1} reveals that estimating the optimal action-value distribution requires a large amount of training if we directly use the system utility as the reward in RL. Therefore, we put forward a new reward-clipping mechanism to prevent the target action-value distribution from greatly changing. Specifically, assuming that there are $T$ thresholds that partition the system utility, we set $T+1$ constants whose values are much smaller than the system utility. Then the system utility can be clipped to these $T+1$ constants that are taken as the rewards in RL. For example, if $T=2$ and the clipping constants are $-\eta$, 0, and $\eta$ ($\eta>0$), then the clipping strategy can be formulated by Eq. \eqref{clippingR}, where $c_1$ and $c_2$ ($c_1 > c_2$) are the manually set thresholds: 
	\begin{equation}
		\label{clippingR}
		r=
		\begin{cases} 
			\eta, & {J(\mathbf{w}, \mathbf{d}) \geq c_{1}}, \\
			0, & {c_{2}<J(\mathbf{w}, \mathbf{d})<c_{1}},   \\
			-\eta, & {J(\mathbf{w}, \mathbf{d}) \leq c_{2}}.
		\end{cases}
	\end{equation}
	However, as $T$ becomes larger, the number of the manually set parameters in the reward-clipping mechanism increases, which makes the parameter setting process more sophisticated. Therefore, we adopt the reward-clipping mechanism defined in Eq. \eqref{clippingR} as an experiment. Note that introducing the reward-clipping mechanism to GAN-DDQN algorithm is effortless, and we only need to apply the reward-clipping mechanism to the system utility before storing the transition tuple in the replay buffer, which is described in line 9 of Algorithm 1. 
	
	\label{Dueling GAN-DDQN}
	\begin{algorithm}
		\label{algorithm2}
		\caption{Dueling GAN-DDQN}
		\begin{algorithmic}[1]
			\STATE Initialize a dueling generator $G$ and a discriminator $D$ with random weights $\bm{\theta}_G$ and $\bm{\theta}_D$ respectively, the number of particles $N$, gradient penalty coefficient $\lambda$, batch size $m$, discount factor $\gamma$, $n_{critic}=5$.
			\STATE Initialize a target dueling generator $\hat{G}$ with weight $\bm{\theta}_{\hat{G}} \gets \bm{\theta}_G$, a replay buffer $\mathcal{B}\gets\varnothing$, the iteration index $t=0$.
			\REPEAT 
			\STATE The agent observes $S_t=s$.
			\STATE The agent samples $\tau \sim U(0, 1)$.
			\STATE The agent feeds $s$ and $\tau$ to network $G$, getting the state-value particles $G_v(s, \tau)$ and each action advantage value $G_{ad}^{(a)}(s, \tau), \forall a \in \mathcal{A}$.
			\STATE The agent calculates $V(s) = \dfrac{1}{N}\sum{G_v(s, \tau)}$.
			\STATE The agent calculates $Q(s, a)=V(s)+G_{ad}^{(a)}(s, \tau), \forall a \in \mathcal{A}$.
			\STATE The agent performs $a^* \gets \arg\max_{a} Q(s_t, a)$.
			\STATE The agent receives the system utility $J$ and observes a new state $S_{t+1}=s^\prime$.
			\STATE The agent performs the reward-clipping with respect to $J$ and gets the reward $r$.
			\STATE The agent stores transition $(s, a^*, r, s^\prime)$ in $\mathcal{B}$.
			\STATE \# \emph{Train network $D$}
			%\REPEAT
			\FOR{$n=1$ to $n_{critic}$}
			\STATE The agent randomly samples $\left\{s, a, r, s^{\prime}\right\}_{i=1}^{m}$ from $\mathcal{B}$.
			\STATE The agent samples $\{\tau\}_{i=1}^{m}$ and $\{\varepsilon\}_{i=1}^{m}$ from $U(0,1)$.
			\STATE The agent gets $y_i = G_v(s_i, \tau_i)$, and $\hat{y}_i = r_i + \gamma\hat{G}_v(s^{\prime}_i, \tau_i)$.
			\STATE The agent sets $\hat{x}_i=\varepsilon_i \hat{y}_i + (1-\varepsilon_i) y_i$.
			\STATE The agent updates the weights $\bm{\theta}_D$ by leveraging gradient descent algorithm to $\frac{1}{m} \sum^{m}_{i=1}\mathcal{L}_i$, where $\mathcal{L}_i=D(y_i) - D(\hat{y}_i) + \lambda \left(\left\|\nabla_{\hat{{x}}_i} D(\hat{{x}}_i)\right\|_{2}-1\right)^{2}$.
			\ENDFOR
			\STATE \# \emph{Train network $G$}
			\STATE The agent randomly samples $\left\{s, a, r, s^{\prime}\right\}_{i=1}^{m}$ from $\mathcal{B}$.
			\STATE The agent samples $\{\tau\}_{i=1}^{m}$ from $U(0,1)$.
			\STATE The agent calculates the estimated action value $Q_i = \dfrac{1}{N}\sum{G_v(s_i, \tau)} + G_{ad}^{(a_i)}(s_i, \tau)$, and the target action value $\hat{Q}_i=r_i + \gamma\dfrac{1}{N}\sum{G_v(s_i^{\prime}, \tau)} + \gamma\max_a G_{ad}^{(a)}(s_i^{\prime}, \tau), \forall a \in \mathcal{A}$.
			\STATE The agent updates the weights $\bm{\theta}_G$ by leveraging gradient descent algorithm to $\frac{1}{m} \sum^{m}_{i=1} [-D(G_v(s_i)) + \frac{1}{2}(\hat{Q}_i-Q_i)^2]$.
			\STATE The agent clones network $G$ to the target network $\hat{G}$ every $C$ iterations by resetting $\bm{\theta}_{\hat{G}} = \bm{\theta}_G$.
			\STATE The iteration index is updated by $t \gets {t+1} $.
			\UNTIL A predefined stopping condition (e.g., the $\frac{1}{m} \sum^{m}_{i=1}\mathcal{L}_i$, the preset number of iterations, etc.) is satisfied.		
		\end{algorithmic}
	\end{algorithm}

	\subsection{Dueling GAN-DDQN}
	The training of GAN-DDQN is not a trivial task since it uses the data yielded from a dynamic environment, and only a tiny portion of the output of the generator is useful for gradient calculation. One intuitive indicating to alleviate the training problem is to carefully adjust the values of GAN-DDQN's hyper-parameters, such as the discount factor $\gamma$, the gradient penalty coefficient $\lambda$, etc. Nevertheless, we plan to make systemic and architectural changes to the generator and the loss function. Particularly, inspired by \cite{qi2019deep}, which uses a specialized dueling Q network to separate the action value into a state-value stream and an advantage stream, we divide the approximation of the action-value distribution into the approximation of the state-value distribution and the approximation of the advantage function for each action. This dueling architecture ignores the trivial variations of the environment and focuses on some crucial states to enhance the stability of DRL algorithms \cite{wang2015dueling}. In addition, in our improved model, namely Dueling GAN-DDQN, the loss function of the discriminator turns to work on the estimated and target state-value distributions. Moreover, the squared TD error is added to the generator's loss as the criterion that measures the distance of the estimated and target action values.
	
	The detailed structure of Dueling GAN-DDQN is presented in Fig. \ref{fig:duelingGANDDQN}(b), and we remarkably highlight the key differences from GAN-DDQN. It can be observed that the significant difference compared with GAN-DDQN is the generator or the dueling generator for the sake of distinguishing. In the dueling generator, after Hadamard product operation, we continue to handle the output using multiple neural layers (the \textit{common layers}). Then, the refined information is separated into two paths, one flowing to a neural network to approximate the state-value distribution, and the other flowing to another neural network to estimate the action advantage function. Accordingly, the dueling generator outputs not only particles from the approximated state-value distribution but also the estimated action advantage values for each action. Note that the discriminator of Dueling GAN-DDQN has the same structure as GAN-DDQN.
	
	Similarly to our analysis of the random variable $Z^\pi_q$, we analyze the random variable $Z^\pi_v(s)$, which denotes the return obtained by following a policy $\pi$ from state $s$. Then we have
	\begin{equation}
		\label{zandv}
		V^{\pi}(s)=\mathbb{E}\left[Z_v^{\pi}(s)\right],
	\end{equation}
	and an analogous distributional Bellman equation for $Z_v$
	\begin{equation}
		\label{belleqv}
		Z_v^{\pi}(s) \stackrel{D}{=} \underset{\underset{s^\prime\sim\mathcal{S}}{a\sim\mathcal{A}}}{\mathbb{E}}[R+\gamma Z_v^{\pi}(s^{\prime})].
	\end{equation}
	It is difficult to find the distributional Bellman optimality operator for $Z_v^{\pi}$. Even worse, Eq. \eqref{belleqv} indicates that the iterative calculation of $Z_v^{\pi}$ requires a reward from every state-action pair, which is a noticeable time-consuming operation. Therefore, we introduce a degraded but simplified method to estimate $Z_v^{\pi}$, which is to minimize the difference between the estimated $Z_v^{\pi}$ and $\mathcal{T}Z_v^{\pi}$ calculated by
	\begin{equation}
		\label{TDv}
		\mathcal{T}Z_v^{\pi} \stackrel{D}{=} r+\gamma Z_v^{\pi}(s^{\prime}),
	\end{equation}
	where $r$ and $s^{\prime}$ are from the transition $(s, a, r, s^\prime)$ sampled from the replay buffer. This degraded approximation may fail to find the optimal state-value distribution, yet it can significantly reduces computation time. In addition, only considering the $1$-Wasserstein loss for the state-value distribution results in the network $G$ weights related to the action advantage function not being trained. %Fortunately, the dueling generator allows us to get the action values by combining the state-value distribution with the action advantages. 
	Therefore, we leverage the TD error to measure the difference of the estimated and the target action values, where the action value is calculated by adding the corresponding action advantage value to the mean of the state-value particles. As a consequence, the ultimate loss function for training Dueling GAN-DDQN is composed of the $1$-Wasserstein distance and the squared TD error, which can be formulated as follows
	\begin{align}
	\label{DuelingDloss}
	\begin{split}
	\mathcal{L}_D &= \underset{\underset{(s, a) \sim \mathcal{B}}{\tau \sim U(0, 1)}}{\mathbb{E}}[D(G_v(s, \tau))]\\
	&- \underset{\underset{(r, s^{\prime}) \sim \mathcal{B}}{\tau \sim U(0, 1)}}{\mathbb{E}}[ D(r+\gamma \hat{G}_v(s^\prime, \tau))]+p(\lambda),
	\end{split} \\
	\label{DuelingGloss}
	\mathcal{L}_G &= -\underset{\underset{(s, a) \sim \mathcal{B}}{\tau \sim U(0, 1)}}{\mathbb{E}}[D(G_v(s, \tau))] + \frac{1}{2}\zeta^{2},
	\end{align}
	where $G_v$ denotes the state-value particles output by the dueling generator, and $\zeta^{2}$ is the squared TD error as defined in Eq \eqref{eq2}. Algorithm 2 and Fig. \ref{fig:duelingGANDDQN}(b) provide the details of Dueling GAN-DDQN.

	\begin{table*}  
		\centering
		\caption{A Brief Summary of Key Settings for Traffic Generation Per Slice}
		\label{tab:traffic} 
		\begin{tabular}{m{3cm} | m{3cm} | m{3cm} | m{3cm} } 
			\toprule[0.8pt] 
			& VoLTE & Video & URLLC  \\  
			\midrule[0.8pt] 
			Bandwidth & \multicolumn{3}{l}{20 MHz}\\
			\hline
			Scheduling & \multicolumn{3}{l}{Round robin per slot (0.5 ms)}\\
			\hline
			Slice Band Adjustment (Q-Value Update) & \multicolumn{3}{l}{1 second (2000 scheduling slots) }\\
			\hline
			Channel & \multicolumn{3}{l}{Rayleigh fading}\\
			\hline 
			User No. (100 in all) & 46 & 46 & 8\\
			\hline
			Distribution of Inter-Arrival Time per User& Uniform [Min = 0, Max = 160ms] & Truncated stationary distribution [Exponential Para = 1.2, Mean = 6 ms, Max = 12.5 ms] & Exponential [Mean = 180 ms] \\
			\hline
			Distribution of Packet Size & Constant (40 Byte) &  Truncated Pareto [Exponential Para = 1.2, Mean = 100 Byte, Max = 250 Byte]  & Variable constant: \{6.4, 12.8, 19.2, 25.6, 32\} KByte or \{0.3, 0.4, 0.5, 0.6, 0.7\} MByte\\
			\hline 
			SLA: Rate & 51 Kbps & 100 Mbps & 10 Mbps \\
			\hline
			SLA: Latency & 10 ms & 10 ms & 1 ms\\
			\bottomrule[1.5pt] 
		\end{tabular}  
	\end{table*} 
	
	\begin{table}  
		\centering
		\caption{The Mapping from Resource Management for Network Slicing to RL Environment}
		\label{tab:envmapping} 
		\begin{tabular}{m{2cm} | m{5cm}} 
			\toprule[0.8pt] 
			RL Environment & Radio Resource Slicing \\  
			\midrule[0.8pt] 
			State & The number of arrived packets in each slice within a specific time window\\
			\hline
			Action & bandwidth allocation to each slice\\
			\hline
			Reward & Clipped weighted sum of SE and SSR in 3 sliced bands\\
			\bottomrule[1.5pt] 
		\end{tabular}  
	\end{table} 
	
	\section{Simulation Results and Numerical Analysis}
	\label{sec:simulation}
	\subsection{Simulation Environment Settings}
	In this part, we verify the performance of GAN-DDQN and Dueling GAN-DDQN in a RAN scenario where there are three types of services (i.e., VoLTE, video, and URLLC) and three corresponding slices in one serving BS, as in \cite{li2018deep}. There exist 100 registered subscribers randomly located within a 40-meter-radius circle surrounding the BS. These subscribers generate standard service traffics as summarized in Table \ref{tab:traffic} based on 3GPP TR 36.814 \cite{3GPPTR36.814} and TS 22.261 \cite{3GPPTS22.261}. The total bandwidth is 10 MHz, and the bandwidth allocation resolution is 1 MHz or 200 KHz. We will show the simulation results for both cases. On the other hand, the packet size of URLLC service has a strong influence on the system utility. For example, it is difficult to meet the latency requirement of URLLC service when the packet size is large, if there is insufficient bandwidth guaranteed for transmission. As a result, SSR degrades, and the system utility is reduced. Therefore, we simulate the network slicing scenario with suitably-sized URLLC packets.
	
	\begin{figure*}
		\centering
		\includegraphics[width=0.8\linewidth]{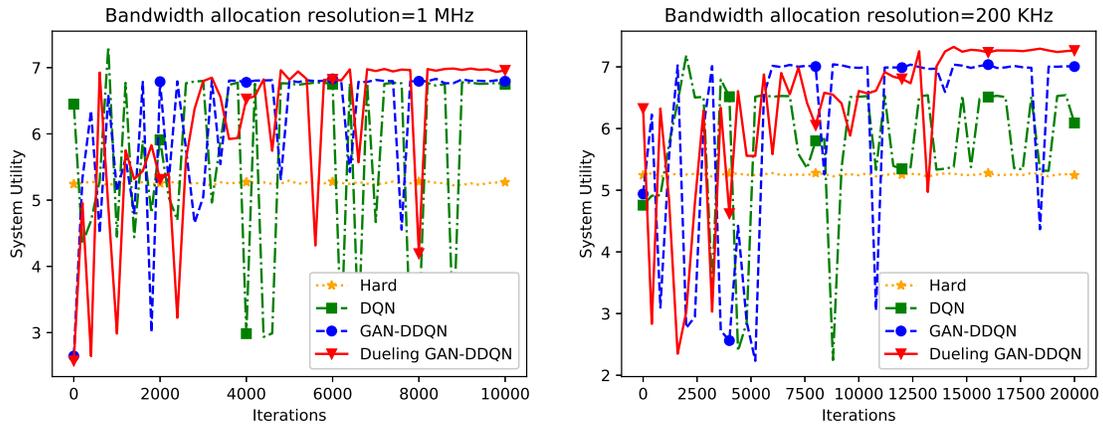}
		\caption{An illustration of performance comparison between different slicing schemes (hard slicing, DQN, GAN-DDQN, Dueling GAN-DDQN).}
		\label{fig:reward_SURLLC}
	\end{figure*} 
	
	\begin{figure*}
		\centering
		\includegraphics[width=1\linewidth]{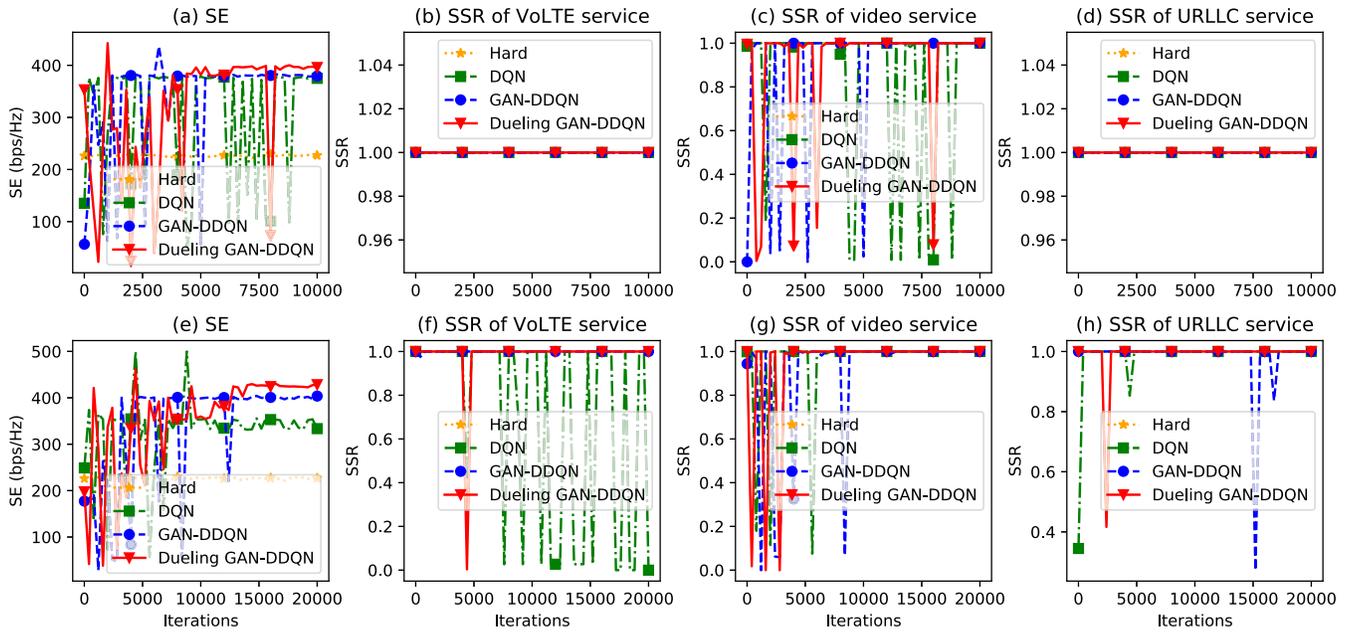}
		\caption{An illustration of SE and SSR in the different cases where the bandwidth allocation resolution is 1 MHz (shown in the top sub-figures) and 200 KHz (shown in the bottom sub-figures).}
		\label{fig:se_qoe}
	\end{figure*}
	
	With the mapping shown in Table \ref{tab:envmapping}, RL algorithms can be used to optimize the system utility (i.e., the weighted sum of SE and SSR). Specifically, we perform round-robin scheduling within each slice at 0.5 ms granularity; that is, we sequentially allocate the bandwidth of each slice to the active users within each slice every 0.5 ms. Besides, we adjust the bandwidth allocation to each slice per second. Therefore, the agent updates its neural networks every second. Considering the update interval of the bandwidth-allocation process is much larger than the service arrival interval, the number of arrived packets (i.e., the state) is rarely zero when updating the agents. Therefore, it is reasonable to ignore the situation of zero bandwidth for any NS. Moreover, this filter setting narrows the range for the action exploration, as well as enhancing the stability of the training process. Meanwhile, this filter setting does affect our main results.

	\subsection{Simulation Results}
	In this part, we show the simulation results of the proposed GAN-DDQN and Dueling GAN-DDQN algorithms, in comparison with the hard slicing method and the standard DQN-based scheme. Hard slicing means that each service is always allocated with $\frac{1}{3}$ of the whole bandwidth (because there are three types of services in total); round-robin scheduling is conducted within each slice. The DQN-based bandwidth allocation scheme was first proposed in \cite{li2018deep}, which directly applied the original DQN algorithm \cite{mnih2015human} to the network slicing scenario. Notably, our previous works in \cite{li2018deep} have demonstrated the classical DQN-driven method is superior to other machine learning methods (e.g., long short-term memory (LSTM)-based prediction-before-allocation method). Therefore, due to the space limitation, we put more emphasis on the performance comparison with the classical DQN in \cite{li2018deep}.

   \subsubsection{Small Packets for URLLC service}
	We first observe the performance of the proposed algorithms within the scenario that the packet size of URLLC service is small. The traffic parameters are shown in Table \ref{tab:traffic}. We consider two cases: the bandwidth allocation resolution is either 1 MHz or 200 KHz. The importance weights in the optimization objective (i.e., Eq \eqref{optimization}) are set to $\alpha=0.01, \bm{\beta}=[1,1,1]$. The values of the clipping parameters $c_{1}$ and $c_{2}$ are determined heuristically\footnote{We first directly regard the system utility as the reward in order to find the range of the system utility, and then try different combinations of the two parameters (i.e., $c_{1} $ and $c_{2}$) to find the suitable values that guarantee both performance and stability.}. In both cases, we set $c_{1}=6.5, c_{2}=4.5$ to clip the system utility according to Eq. \eqref{clippingR}, where $\eta$ is fixed at 1. The experimental evaluation of the reward-clipping setting is investigated hereinafter. Fig. \ref{fig:reward_SURLLC} depicts the variations of the system utility with respect to the iteration index. The left part of Fig. \ref{fig:reward_SURLLC} shows that when the bandwidth allocation resolution is 1 MHz, the three RL-based algorithms perform similarly, but Dueling GAN-DDQN is slightly better; DQN is the most erratic in training. The right part of Fig. \ref{fig:reward_SURLLC} illustrates that when the bandwidth allocation resolution becomes 200 KHz, GAN-DDQN and Dueling GAN-DDQN expand the gap to DQN, which demonstrates the performance improvement coming from distributional RL by the characterization of the action-value or state-value distributions. 
	\begin{figure*}
		\centering
		\includegraphics[width=0.8\linewidth]{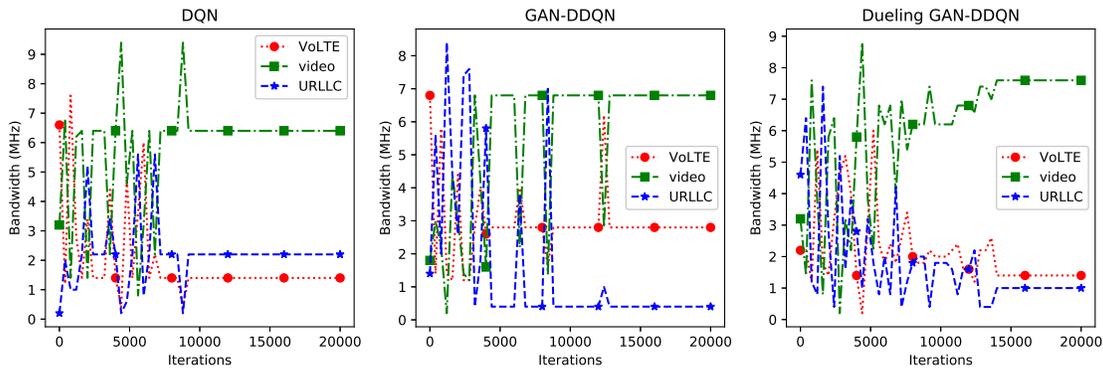}
		\caption{An illustration of bandwidth allocation schemes in the case where the bandwidth allocation resolution is 200 KHz.}
		\label{fig:policy}
	\end{figure*}
	\begin{figure}
		\centering
		\includegraphics[width=0.8\linewidth]{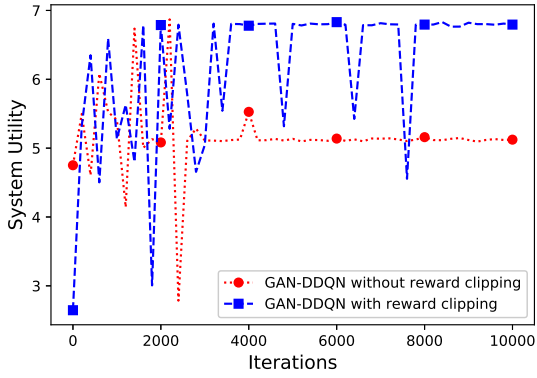}
		\caption{Comparison of GAN-DDQN with and without reward clipping.}
		\label{fig:reward-clipping}
	\end{figure}
	It's worth noting that Dueling GAN-DDQN improves visibly over GAN-DDQN in both performance and stability, consistent with our previous discussion. Furthermore, it can be observed in Fig. \ref{fig:reward_SURLLC} that the system utility obtained by the three RL-based algorithms is significantly greater than that for the hard slicing scheme. The reason for this lies in that the RL-based algorithms can dynamically and reasonably manage bandwidth resources, thereby avoiding wasted resources and improving resources utilization. Moreover, the utilization of the bandwidth resource further gets improved when we slice the bandwidth more finely. Fig. \ref{fig:reward_SURLLC} shows that the system utility obtained by the two GAN-based algorithms, especially Dueling GAN-DDQN, becomes significantly larger when the bandwidth allocation resolution changes to 200 KHz from 1 MHz, but that the performance of DQN is slightly degraded.

	Fig. \ref{fig:se_qoe} presents the variations of SE and SSR with respect to the iteration index for both bandwidth allocation resolution settings (i.e., 1 MHz and 200 KHz). It can be observed from Fig. \ref{fig:se_qoe} that SE curves are basically consistent with the system utility curves. However, the SSR curves of the three algorithms for the NSs show different patterns. When the bandwidth allocation resolution is 1 MHz, the SSRs for both VoLTE and URLLC services reach 100\% with iterative training. Nevertheless, for the SSR of video service, GAN-DDQN and Dueling GAN-DDQN basically converge to 100\% after 5000 iterations, while DQN shows no obvious signs of convergence. When the bandwidth allocation resolution is 200 KHz, it can be observed that GAN-DDQN and Dueling GAN-DDQN, by and large, realize 100\% of SSR for all three services by the end of training, but DQN shows extreme instability for VoLTE service. Note that the unusual sudden performance drop late in training is caused by the tiny nonzero exploration rate in the $\epsilon$-greedy exploration strategy.
	
	\begin{figure}
		\centering
		\includegraphics[width=0.8\linewidth]{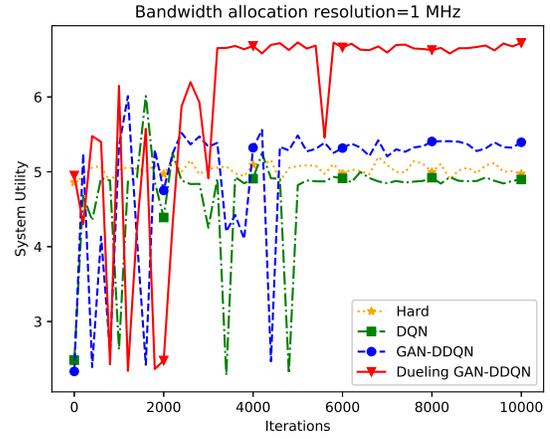}
		\caption{An illustration of performance comparison between different slicing schemes in the case where the packets of URLLC service are large and the bandwidth allocation resolution is 1 MHz.}
		\label{fig:reward_LURLLC}
	\end{figure}

	In Fig. \ref{fig:policy}, we illustrate the policy learned by the three algorithms when the bandwidth allocation resolution is 200 KHz. It can be observed that all three algorithms converge after 15000 iterations. However, there are some differences between the learned bandwidth allocation policies. The DQN agent allocates the least bandwidth to VoLTE service and keeps it unchanged; as a result, SSR of VoLTE service does not always reach 100\%. Between the GAN-DDQN agent and the Dueling GAN-DDQN agent, the latter behaves more intelligently, which is prominently manifested in the fact that it maximizes the bandwidth allocated to video service and reduces the bandwidth allocated to the other two services while meeting the SLA. The Dueling GAN-DDQN agent provides as much bandwidth as possible to satisfy the SLA of video service which is requested frequently and bandwidth-consuming, thus improving the SE. Besides, Dueling GAN-DDQN agent provides a policy to better balance the demands of VoLTE and URLLC services that are relatively rarely requested.
	
	We next investigate the impact of the reward-clipping mechanism. Fig. \ref{fig:reward-clipping} shows the differences in system utility during the iterative learning of GAN-DDQN with and without the reward clipping when the bandwidth allocation resolution is 1 MHz. When there is no reward clipping, GAN-DDQN directly takes the system utility as the reward. Note that the values of system utility, although they fluctuate, are much larger than the clipping constants set manually in the reward-clipping mechanism. As a result, if the system utility is directly used as the reward, the target action-value distribution might vary significantly, making it difficult for GAN-DDQN to converge. Therefore, GAN-DDQN without reward clipping requires more training steps to converge from one equilibrium to a new one. It can be observed from Fig. \ref{fig:reward-clipping} that the GAN-DDQN performs significantly better with reward clipping than without. The simulation results verify the effectiveness of GAN-DDQN together with reward clipping.

	\subsubsection{Large Packets for URLLC service}
	\begin{figure}
		\centering
		\includegraphics[width=1.\linewidth]{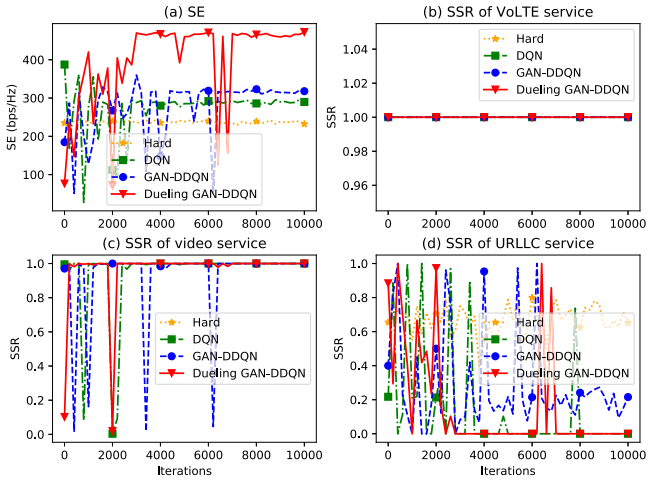}
		\caption{An illustration of SE and SSR in the case where the bandwidth allocation resolution is 1 MHz.}
		\label{fig:se_qoe_LURLLC}
	\end{figure}
		\begin{figure}
		\centering
		\includegraphics[width=1.\linewidth]{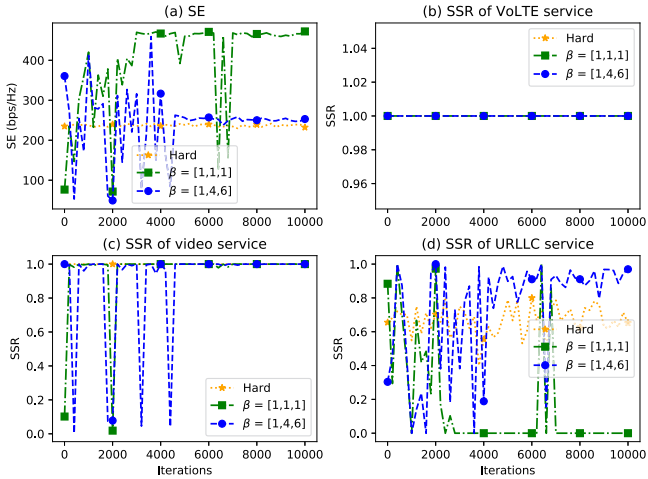}
		\caption{An illustration of SE and SSR achieved by Dueling GAN-DDQN when the bandwidth allocation resolution is 1 MHz and the importance weights of SSR are $[1, 1, 1]$ and $[1, 4, 6]$, respectively.}
		\label{fig:se_qoe_LURLLC_146}
	\end{figure}
	In this part, we consider the case where the packet size of URLLC service is evenly sampled from $\{0.3, 0.4, 0.5, 0.6, 0.7\}$ MByte, which gives a considerably larger packet size than we just analyzed and requires more bandwidth to guarantee meeting the SLA of URLLC service. In the case where bandwidth allocation resolution is 1 MHz, we set $c_{1}=5.7, c_{2}=3$ to clip the system utility according to Eq. \eqref{clippingR}, where $\eta$ is fixed to 1.
	Fig. \ref{fig:reward_LURLLC} shows the performance of each slice algorithm, from which it can be observed that Dueling GAN-DDQN is way ahead of the others in terms of system utility. However, quite unexpectedly, DQN performs poorly, worse even than hard slicing scheme. Fig. \ref{fig:se_qoe_LURLLC} reveals the details of SE and SSR, from which we can find that Dueling GAN-DDQN agent learned a policy that maximizes system utility by sacrificing the SSR of URLLC service in exchange for higher SE. The reason for this is that when SSR is equally important to all three services (i.e., $\bm{\beta}=[1,1,1]$), it is challenging for URLLC service to satisfy its SLA given the large transmission volume and the strictly low latency requirement. Therefore, we further investigate the situation in which URLLC service is more concerned while keeping video service dominating VoLTE service, which is reflected in the value of $\bm{\beta}$ changing to $[1, 4, 6]$ from $[1, 1, 1]$.
	Fig. \ref{fig:se_qoe_LURLLC_146} presents the results for these different $\bm{\beta}$ settings and demonstrates that the adjusted importance weight makes the SLA of URLLC service well guaranteed; meanwhile, the SLA of the other two services can be 100\% satisfied as well. However, the requests for URLLC service are scarce in a timeslot---in our simulation, it is the interval between two updates of the agent, defined as 1 second---while the agent has to allocate more bandwidth to URLLC slice to guarantee conformity to SLA until the next update, which wastes the bandwidth to some extent and thus leads to the decrease of SE. There are two lessons that we learn from these simulation results: (a) it is non-trivial to optimize multiple conflicting objectives, even when using cutting-edge RL algorithms; (b) shortening the interval between successive bandwidth allocations may improve performance but it also increases computational costs and raises issues of stability due to more drastic changes in demand.
	
	\section{Conclusion}
	In this paper, we have investigated the combination of deep distributional RL and GAN and proposed GAN-DDQN to learn the optimum solution for demand-aware resource management in network slicing. In particular, we have applied GAN to approximate the action-value distribution, so as to avoid the negative impact of randomness and noise on the reward and grasp much more details therein than the conventional DQN. We have also designed a new update procedure that combines the advantages offered by distributional RL with the training algorithm of WGAN-GP. Furthermore, we have adopted the reward-clipping scheme to enhance the training stability of GAN-DDQN. Besides, we have introduced the dueling structure to the generator (i.e., Dueling GAN-DDQN), so as to separate the state-value distribution and the action advantage function from the action-value distribution and thus avoid the inherent training problem of GAN-DDQN. Extensive simulations have demonstrated the effectiveness of GAN-DDQN and Dueling GAN-DDQN with superior performance over the classical DQN algorithm. In the future, we will try to further improve the GAN-DDQN mechanism under various scenarios with multiple-metric constraints as well as non-stationary traffic demands.
	
	\begin{appendix}
		\section*{The proof of Theorem 1}
		Before the proof of Theorem \ref{therorem1}, we give the following lemmas:
		\newtheorem{lemma}{Lemma}
		\begin{lemma}
			\label{lemma1}
			\textit{Because $v(\theta, \psi)=0$ if and only if $(\theta, \psi)=(\xi, 0)$, the unique Nash-equilibrium point of the training objective in Eq. \eqref{trainingobj} is given by $\theta=\xi, \psi=0$.}
		\end{lemma}
		\begin{lemma}
			\label{lemma2}
			\textit{The distance between the optimal boundaries of Dirac-WGAN-GP on $\mathcal{D}_1$ and $\mathcal{D}_2$ is $\delta$}.
		\end{lemma}
	
		\newenvironment{proof}{{\indent\it Proof} : }{\hfill $\blacksquare$}
		\begin{proof}
			Dirac-WGAN-GP consists of a generator with parameter $\theta$ and a linear discriminator $D_{\psi}(x)=\psi \cdot x$, where the generator outputs a Dirac distribution centralized $\theta$ (i.e., $\delta_{\theta}$). Whilst the real data distribution is given by a Dirac distribution concentrated at $\xi$ (i.e., $\delta_{\xi}$). Therefore, the training objective of Dirac-WGAN-GP is given by 
			\begin{equation}
				\label{trainingobj}
				L(\theta, \psi)=\psi \theta-\xi\psi
			\end{equation}
			and the gradient penalty proposed in \cite{gulrajani2017improved} is given by
			\begin{equation}
				\begin{aligned}
				p(\psi) &=\frac{\lambda}{2} \mathbb{E}_{\hat{x}}\left(\left\|\nabla_{\hat{x}} D_{\psi}(\hat{x})\right\|-1\right)^{2} \\
				&=\frac{\lambda}{2}\left(|\psi|-1\right)^{2}
				\end{aligned}
			\end{equation}
			
			Inspired by \cite{mescheder2018gan}, we use gradient vector field to analyze convergence, which is defined as follow
			\begin{equation}
			v(\theta, \psi) :=\left( \begin{array}{c}{-\nabla_{\theta} L(\theta, \psi)} \\ {\nabla_{\psi} L(\theta, \psi)}\end{array}\right)
			\end{equation}
			For Dirac-WGAN-GP, the corresponding gradient vector field is given by
			\begin{equation}
			\label{gradvecfield}
			v(\theta, \psi)=\left( \begin{array}{c}{-\psi} \\ {\theta- \xi+ \operatorname{sign}(\psi) \lambda\left(|\psi|-1\right)}\end{array}\right)
			\end{equation}
			where $\operatorname{sign}(\cdot)$ denotes the signum function and we have Lemma \ref{lemma1}.
			
			Assume that the iteration $(\theta_{k}, \psi_{k})$ converges towards the equilibrium point $(\xi, 0)$ but $\left(\theta_{k}, \psi_{k}\right) \neq(\xi,0)$ for all $k \in \mathbb{N}$, which implies that $v(\theta_{k}, \psi_{k})\approx 0$ and thus we have
			\begin{equation}
				-\psi_{k} \approx \theta_{k}- \xi+ \operatorname{sign}(\psi_{k}) \lambda\left(|\psi_{k}|-1\right)
			\end{equation}
			in other words,
			\begin{equation}
					\theta_{k} \approx -\psi_{k}+\xi-\operatorname{sign}(\psi_{k})\lambda(|\psi_{k}|-1)
			\end{equation}
			Then, we can get the update amount of parameter $\theta$ after the ($k+1$)th training as follow 
			\begin{equation}
			\begin{aligned}
			&\left|\theta_{k+1}-\theta_{k}\right| \\&\approx h\left|-\psi_{k}+\xi-\operatorname{sign}\left(\psi_{k}\right)\lambda(|\psi_{k}|-1)-\theta_{k}\right|\\
			&\approx h\left|-(\lambda+1)\psi_{k}+(\xi-\theta_{k})+\operatorname{sign}\left(\psi_{k}\right)\lambda\right|
			\end{aligned}
			\end{equation}
			Therefore, we have  $\lim _{k \rightarrow \infty}\left|\theta_{k+1}-\theta_{k}\right|=h\lambda$,
			which shows that Dirac-WGAN-GP cannot converge to the equilibrium point, and the value of generator's parameter will finally oscillate between $\xi-\frac{h\lambda}{2}$ and $\xi+\frac{h\lambda}{2}$. 
			
			Assume that $\mathcal{D}_1$ and $\mathcal{D}_2$ are two different real data, which are the Dirac distributions concentrated at $\xi_1$ and $\xi_2$, respectively. Let $\delta =|\xi_{1}-\xi_{2}|$, which indicates the statistic distance between $\mathcal{D}_1$ and $\mathcal{D}_2$. Note that usually $h\lambda$ is two or three orders of magnitude smaller than $\delta$, which implies that the optimal boundaries is rarely overlapped, thus further training is required when the real data varies. With the constant learning rate,  it is easy to deduce from Lemma \ref{lemma2} that the larger the $\delta$ is, the more training steps are required for the Dirac-WGAN-GP to reach the new optimal boundary. Finally, based on Lemma \ref{lemma1} and \ref{lemma2}, we obtain the proof of Theorem 1. 
		\end{proof}
			 
	\end{appendix}

\bibliographystyle{unsrt}

\end{document}